\documentclass{article}
\usepackage{spconf,amsmath,graphicx}
\usepackage{cite}
\usepackage{epsfig,dblfloatfix}

\usepackage{amsfonts,amsthm}		
\usepackage{algorithm, algorithmic}

\title{Level Set Estimation from Compressive Measurements using\\
Box Constrained Total Variation Regularization}
\name{Akshay Soni and Jarvis Haupt}
\address{University of Minnesota, Twin Cities\\
Department of Electrical and Computer Engineering\\
Minneapolis, Minnesota USA 55455\\
e-mail: {\tt \{sonix022,jdhaupt\}@umn.edu}}
\begin{document}
\ninept
\maketitle
\begin{abstract}
Estimating the level set of a signal from measurements is a task that arises in a variety of fields, including medical imaging, astronomy, and digital elevation mapping.  Motivated by scenarios where accurate and complete measurements of the signal may not available, we examine here a simple procedure for estimating the level set of a signal from highly incomplete measurements, which may additionally be corrupted by additive noise. The proposed procedure is based on box-constrained Total Variation (TV) regularization.  We demonstrate the performance of our approach, relative to existing state-of-the-art techniques for level set estimation from compressive measurements, via several simulation examples.
\end{abstract}



%

\section{Introduction}

Let $x \in \mathbb{R}^{p}$ represent our signal of interest. A $\gamma$-\emph{level set} of $x$ is defined as the set of locations where the value of the signal $x$ exceeds some specified threshold $\gamma$; i.e. $S^* = S^*(\gamma) = \{j :x(j) \geq \gamma\},~j = 1,...,~p$. Identification of level sets plays a crucial role in a variety of applications such as medical imaging where, for example, level sets can indicate presence of pathologically significant features such as tumors \cite{KWR2011, SD2007}. 

If $x$ is known exactly, the level set estimation task is of course trivial.  Here, our focus is on settings where $x$ itself may not be directly available; instead, we only have access to linear measurements of $x$, which may be noisy and/or incomplete.  Consider, for example, the case where complete noisy measurements of the signal of interest are available, such that measurements are of the form
\begin{equation}
\label{eq:meas}
y = x + n,
\end{equation}
where $y \in \mathbb{R}^{p}$ and $n \in \mathbb{R}^p$ is an i.i.d. noise vector.  In this case, a simple approach to level set estimation would entail coordinate-wise thresholding, estimating from the noisy measurements the regions where $x$ exceeds the specified threshold $\gamma$. 

In more interesting settings, measurements may come in the form of noisy linear projections of the unknown signal,
\begin{equation}
\label{eq:meas1}
y = Ax + n,
\end{equation}
where $A \in \mathbb{R}^{k \times p}$ is a projection matrix and $n \in \mathbb{R}^k$ is additive i.i.d. noise. Such models describe, for example, the measurements obtained in magnetic resonance imaging (MRI) applications, which correspond to samples of the Fourier-domain representation of the signal; similar models describe observations obtained in tomographic imaging applications.  

Generally speaking, the condition $k<p$ describes settings where the number of measurements is less than the ambient dimension of the signal being acquired, a condition that may be imposed, for example, by sampling strategies designed to adhere to physical resource constraints.  Problems involving recovery of high-dimensional signals from undersampled data comprise the essential focus of current research into compressive sensing (CS) (see, eg, \cite{Donoho06, Candes}), where more ``exotic'' measurement operators, such as (pseudo-)random projections, have been examined extensively.  The essential goal in CS is typically to recover (or estimate) the unknown signal from a reduced number of measurements; here, we examine the problem of estimating a level set of $x$ from compressive measurements.

\subsection{Prior Work}

Recent work \cite {KWR2011} initiated the study of level set estimation from compressive measurements, and proposed a strategy for estimating the level set directly from the compressive measurements without performing the reconstruction step.  The approach outlined in \cite{KWR2011} comprises an extension of the level set estimation technique developed in \cite{WN2007}, which was designed for the setting where measurements are complete, but noisy, as described by the model \eqref{eq:meas}.  

The approach proposed and analyzed in \cite{WN2007} is based on a complexity-regularized level set estimator of the form
\begin{equation}
\hat{S} = \arg \underset{S \in \mathcal{S}}{\min}~\hat{R}(S) +\alpha \Phi(S),
\end{equation}
where $\mathcal{S}$ is a class of candidate estimates, $\hat{R}(S)$ is an empirical measure of the estimator risk based on $p$ noisy measurements of the signal, given by
\begin{equation}\label{eqn:empr}
\hat{R}(S) = \frac{1}{p} \sum_i (\gamma - y_i)\left[ \mathbb{I}_{\{i \in S\}} - \mathbb{I}_{\{i \notin S\}} \right],
\end{equation}
where $\mathbb{I}_{\{E\}} = 1$ if event $E$ is true and $0$ otherwise, and $\Phi(S)$ is a carefully designed \emph{tree-based} complexity regularization term which penalizes improbable level sets based proportional to their depth in a recursive dyadic representation of the signal domain. Here, $\alpha>0$ is a parameter that controls the relative influence of each term in the overall cost function. The empirical risk function $\hat{R}(S)$ is devised as a surrogate for the true excess risk of a candidate level set $S$ which is defined as 
\begin{equation}\label{eqn:risk}
\mathcal{E}(S)=\frac{1}{p} \sum_{i \in \triangle(S^*, S)}|\gamma - x_i|
\end{equation}
where $\triangle(S^*, S) = \{i \in (S^* \cap \bar{S}) \cup (\bar{S}^* \cap S)\}$ denotes the symmetric difference, and $\bar{S}$ is the complement of $S$. The essential idea behind this formulation is that estimates having small excess risk are likely close to the true level set, and while \eqref{eqn:risk} cannot be evaluated directly from data (since $S^*$ is unknown), the empirical surrogate \eqref{eqn:empr} can. 

Now, the approach in \cite{KWR2011} is concerned with estimating level sets from compressive measurements, and to that end, proceeds in a manner similar in spirit to the approach of \cite{WN2007}, with an additional ``pre-processing'' step.  Given the measured data $y$, the authors of \cite{KWR2011} suggest forming the ``proxy'' observations
\begin{equation}
\label{proxy}
z = A^{T}y = x + \underbrace{{(A^{T}A - I)x + A^{T}n}}_{\tilde{n}},
\end{equation}
which take the form of a signal plus signal-dependent-noise $\tilde{n}$. The procedure of \cite{WN2007} is then applied to the proxy observations $z$, rather than $y$.  The analysis in \cite{KWR2011} comprises a careful treatment of the signal-dependent noise term $\tilde{n}$, under the framework proposed in \cite{WN2007}.  We refer the reader to \cite{WN2007, KWR2011} for further details and analysis of these existing level set estimation approaches.

\subsection{Contributions}

In this paper we examine an alternative approach, and demonstrate that level sets can be estimated from compressive measurements quickly and accurately using estimation-based techniques. Our method entails solving an optimization problem with total variation (TV) regularization, subject to additional constraints on the solution set.  Our main contribution here is to demonstrate the effectiveness of this estimation-based approach to compressive level set estimation relative to a simple thresholding-based approach, as well as the state-of-the-art approach in \cite{KWR2011}. 

The remainder of the paper is organized as follows. In Section \ref{our approach}, we discuss the optimization problem to be solved for level set estimation, and briefly describe an algorithm for solving the TV regularization problem that is based on the Fast Iterative Shrinkage and Thresholding Algorithm (FISTA) \cite{Beck2009, BT2009}. We evaluate the performance of our approach via simulation, and these results are presented in Section \ref{exp}.  Finally, conclusions and directions for future work are discussed in Section \ref{conc}.

\section{TV Regularization for Level Set Estimation}
\label{our approach}
In this section, we describe an estimation-based algorithm for level set estimation using total variation (TV)  regularization. TV regularization (proposed initially in \cite{Osher}) is now a standard approach in many image denoising and deblurring problems, and fast algorithms have been developed for solving TV minimization problems (see, eg., \cite{BT2009}). Suppose we collect noisy projection measurements of (a vectorized version of) the image of interest, according to the model \eqref{eq:meas1}.  As above we denote by $x\in\mathbb{R}^p$ the signal being acquired. Now, suppose $p=mn$ for some integers $m,n>0$, and let us denote by $X$ the reshaped $m\times n$ image whose vectorized representation is $x$. The discrete penalized version of the TV minimization problem we consider here consists of solving a convex minimization problem of the form, 
\begin{equation}\label{eq:min}
\hat{X} = \mathrm{arg}~\underset{Z}{\rm{min}}~~ \frac{1}{2}\|Az - y\|_2^2 + \alpha \|Z\|_{\rm{TV}},
\end{equation}
where $A \in \mathbb{R}^{k \times p}$ is the linear measurement operator, $z \in \mathbb{R}^{p}$ is the vectorized representation of $Z \in \mathbb{R}^{m \times n}$, and $y \in \mathbb{R}^k$. Here, $\|\cdot\|_{TV}$ could be either the isotropic TV function, given by
\begin{eqnarray}
\nonumber
\lefteqn{\|X\|_{\rm TV, iso}}&&\\
\nonumber&= &\sum_{i=1}^{m-1}\sum_{j=1}^{n-1}{\sqrt{\left(X_{i,j} - X_{i+1,j}\right)^2 + \left(X_{i,j} - X_{i,j+1}\right)^2}} \\
&+& \sum_{i=1}^{m-1}|X_{i,n} - X_{i+1,n}| + \sum_{j=1}^{n-1}|X_{m,j} - X_{m,j+1}|
\end{eqnarray}
or the $\ell_1-$based, anisotropic TV, which is defined by
\begin{eqnarray}
\nonumber
\|X\|_{{\rm TV},\ell_1} &= &\sum_{i=1}^{m-1}\sum_{j=1}^{n-1}{\{|X_{i,j} - X_{i+1,j}| + |X_{i,j} - X_{i,j+1}|\}} \\
\nonumber
&+& \sum_{i=1}^{m-1}|X_{i,n} - X_{i+1,n}| + \sum_{j=1}^{n-1}|X_{m,j} - X_{m,j+1}|.
\end{eqnarray}
The regularization parameter $\alpha > 0$ provides a tradeoff between fidelity to measurements and complexity of the solution, as quantified by the TV norm.

\begin{algorithm}[t]
\caption{FISTA: Level Set Estimation}
\textbf{Input}: $L = \lambda_{\rm max}(A^TA)$; \ $\alpha>0$; \ $l, u\in\mathbb{R} \ (l<u)$\\
\textbf{Initialize}: $\rho =1/L$; \ $t^1 = 1$; $x^0 = r^1 = 0$
\begin{algorithmic}
\WHILE{$\mbox{not converged}$}
\STATE $1: x_g = r^k - \rho\nabla f(r^k)$
\STATE $2: X_{g} = \mathrm{reshape}(x_g, m, n) $
\STATE $3: X_k = \mathrm{prox}_{\rho}(\alpha \|X\|_{\mathrm{TV}})(X_g)$
\STATE $4: x_k = \mathrm{reshape}(X_k, p, 1) $
\STATE $5: x^k = \mathrm{project}(x_k,l,u)$
\STATE $6: t^{k+1} = \left(1+\sqrt{1+4(t^k)^2}\right)/2$
\STATE $7: r^{k+1} = x^k + \left((t^k -1)/t^{k+1}\right)(x^k - x^{k-1})$
\ENDWHILE
\end{algorithmic}

\end{algorithm}

We pose our level set estimation approach in terms of the solution of a box-constrained TV minimization
\begin{eqnarray} \label{eq:level1}
\nonumber
\hat{X} &=& \mathrm{arg}~\underset{z}{\rm{min}}~~ \frac{1}{2}\|Az - y\|_2^2 + \lambda \|Z\|_{\rm{TV}},\\ ~~\mbox{such that}~&l&\leq z_i \leq u,~i = 1,...,p,
\end{eqnarray}
where $l$ and $u$ are the upper and lower values every element of the solution $x$ can take.  In general, $[l,u]$ may describe the range of allowable pixel amplitudes; for $\gamma$-level set estimation, we choose $l$ to be just slightly less than $\gamma$, since for this task we are ultimately uninterested in those pixels whose values are less than $\gamma$.

\begin{figure}[t]
\begin{center}
\begin{tabular}{cc}
\epsfig{file=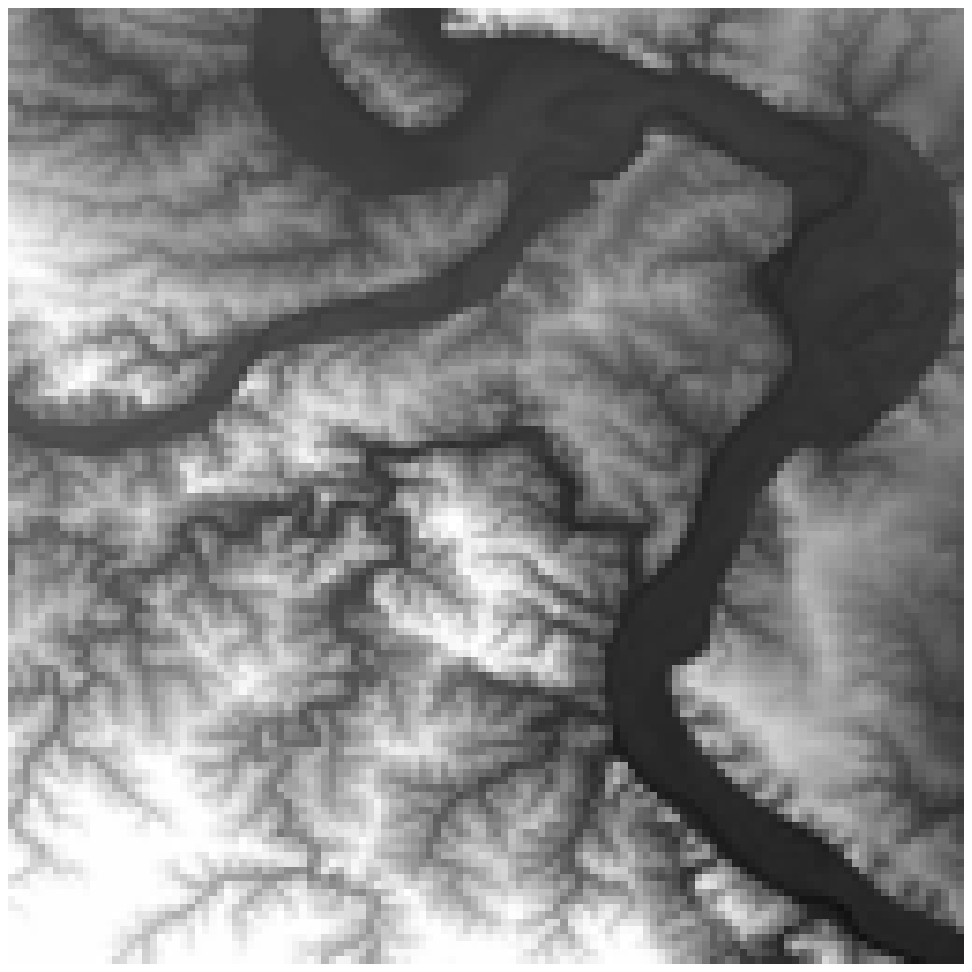,width=0.3\linewidth,clip=} &
\epsfig{file=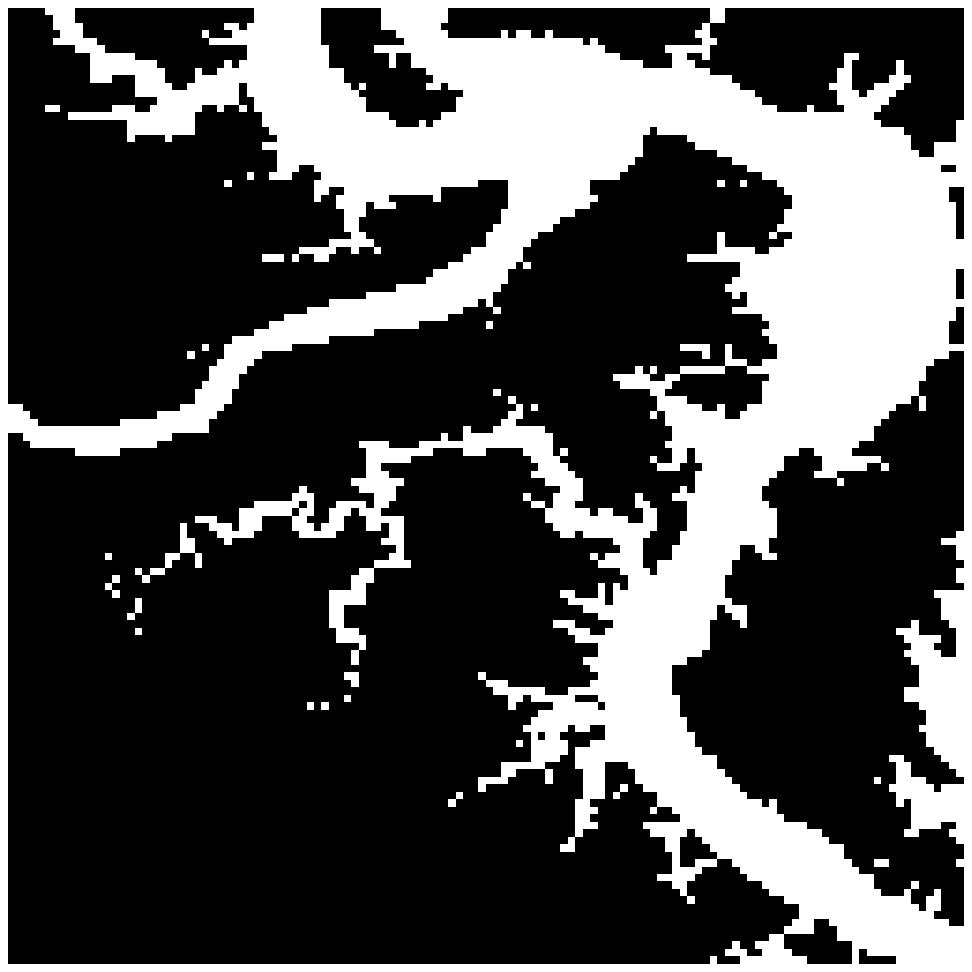,width=0.3\linewidth,clip=} \\
(a) & (b)\\
\epsfig{file=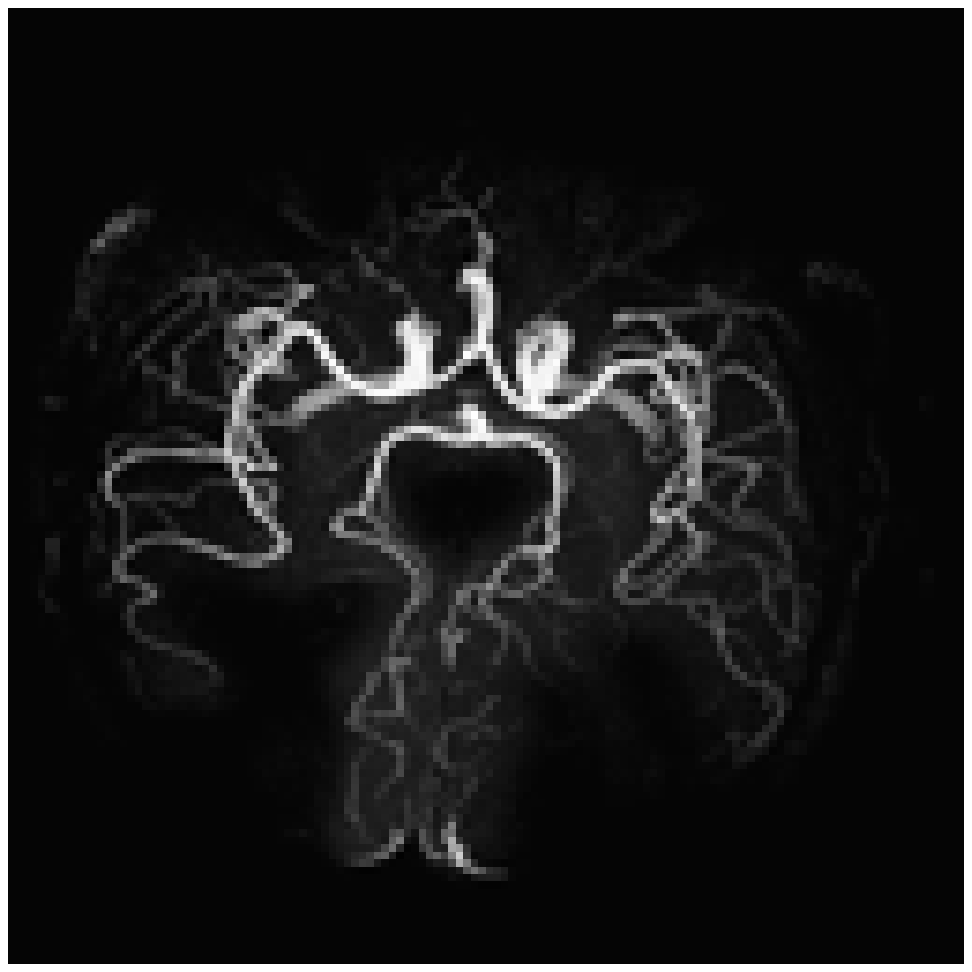,width=0.3\linewidth,clip=} &
\epsfig{file=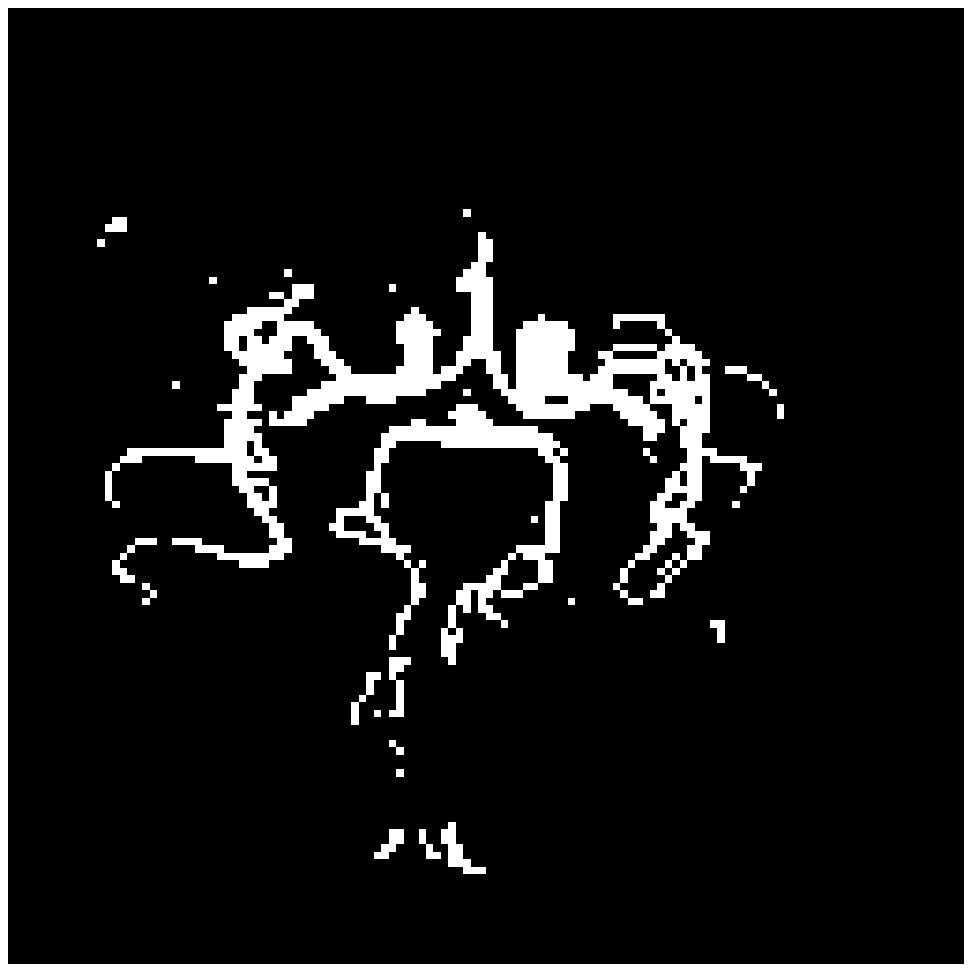,width=0.3\linewidth,clip=} \\
(c) & (d)\\
\end{tabular}
\end{center}
\caption{Sample grayscale images and their corresponding level sets.  Panel (a) depicts an elevation map of St. Louis
with pixel intensities in $[0,255]$, and panel (b) depicts its $\gamma=70$ level set. The image in panel (c) is a magnetic resonance angiography image, also having pixel intensities in $[0,255]$, and its $\gamma=60$ level set is shown in panel $(d)$.}
\label{res1}
\end{figure}

\sloppypar Algorithm 1 describes the procedure that we employ here, which is based on the Fast Iterative Shrinkage and Thresholding Algorithm FISTA \cite{Beck2009, BT2009}. Our optimization takes the general form
\begin{equation}
\min_x f(x) + g(x),
\end{equation}
where
$f(x) = \frac{1}{2}\|Ax - y\|_2^2$ and $g(X) = \alpha \|X\|_{\rm{TV}} + \delta_\mathcal{C}(X)$, and $\delta_\mathcal{C}$ is the indicator function on $\mathcal{C}$.  The efficiency of our FISTA-based approach relies on us being able to quickly obtain the quantity $X_k = \mathrm{prox}_{\rho}(\alpha \|X\|_{\mathrm{TV}})(X_g)$; in general, for a continuous convex function $g(x)$ and $\rho>0$,
\begin{equation}
\mathrm{prox}_{\rho}(g)(x) := \mathrm{arg}~\underset{u}{\mathrm{min}}\left\{ g(u) + \frac{1}{2\rho}\|u - x\|_2^2\right\}.
\end{equation}

\noindent Here, we solve this by using the fast gradient based algorithm described in \cite{BT2009}. The function $\nabla f(x)$ denotes the gradient of the function $f$ at the point $x$, which here is simply given by $\nabla f(x) = A^T(Ax - y)$. The function $\mathrm{project}(x,l,u)$ is given by
\begin{equation}
\mathrm{project}(x,l,u) = \left \{ 
\begin{array}{ll}
x ~~~~~\mathrm{if} ~l \leq x \leq u,\\
l ~~~~~~\mathrm{if} ~x < l,\\
u ~~~~~\mathrm{if} ~x > u.
\end{array} \right.
\end{equation}
where $l$ and $u$ are as described above. The ${\rm reshape}(x,m,n)$ step simply takes a vector/matrix and reshapes it to dimension $m \times n$.  Standard termination criteria can be specified (eg., terminate when the difference between successive estimates is sufficiently small).
\section{Experimental Results}
\label{exp}

%
%
%

We performed level set estimation experiments with two different test images\footnote{The St. Louis elevation map image in Fig.~\ref{res1}(a) is available at {\tt www.usgs.gov/features/lewisandclark/Mapping.html}; the magnetic resonance angiography image in Fig.~\ref{res1}(c) is available at 
{\tt en.wikipedia.org/wiki/Magnetic\_resonance\_angiography}.}, shown in Fig.~\ref{res1}(a) and (c).  Measurements of a particular $128\times 128$ test image $X$ were obtained via the noisy linear model $y = Ax + n$, where $x\in\mathbb{R}^p$ ($p=128^2$) is the vectorized representation of the image $X$, and $A \in \mathbb{R}^{k\times p}$.  The images used for experiments are in $8$-bit binary format, meaning that their pixels take integer values in $[0,255]$.  Here the entries of $A$ are generated as i.i.d.~realizations of $\mathcal{N}(0, 1/k)$ random variables, and the additive noise components of $y$ are i.i.d. $\mathcal{N}(0, \sigma^2)$.  We consider several settings, corresponding to several different choices of the number of measurements $k$ and noise standard deviation $\sigma$.  

We compare the approach outlined here (using the isotropic TV norm) with the state of the art approach in \cite{KWR2011}. As discussed above, for the proposed approach we choose $l$ slightly smaller than the target level (specifically, $l=\gamma-5$ here) and choose $u=255$.  In addition, both our approach, as well as the method in \cite{KWR2011}, are dependent on choice of an algorithmic parameter $\alpha>0$.  Following the evaluation methodology described in \cite{KWR2011}, we choose these parameters clairvoyantly, and display the results obtained using the parameters that result in the minimum excess risk, as defined in \eqref{eqn:risk}.

\begin{figure}[t]
\begin{center}

\epsfig{file=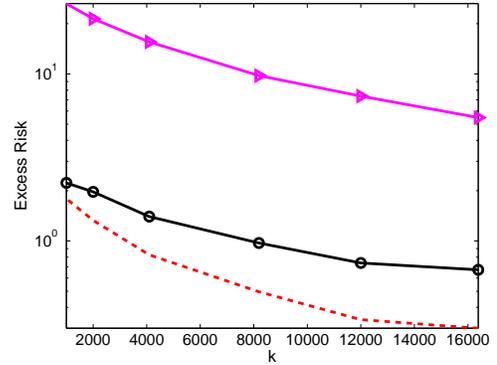,width=0.75\linewidth,clip=}

\end{center}
\caption{Excess risk of image from Fig.\ref{res1}(a) where the $--$ denotes our approach proposed here, -$\circ$- is the approach from \cite{KWR2011} and -$\triangleright$- is the simple thresholding-based approach in a noise-free ($\sigma=0$) setting.}
\label{res2}
\end{figure}

\begin{figure*}[!ht]
\begin{center}
\begin{tabular}{cccccc}
\epsfig{file=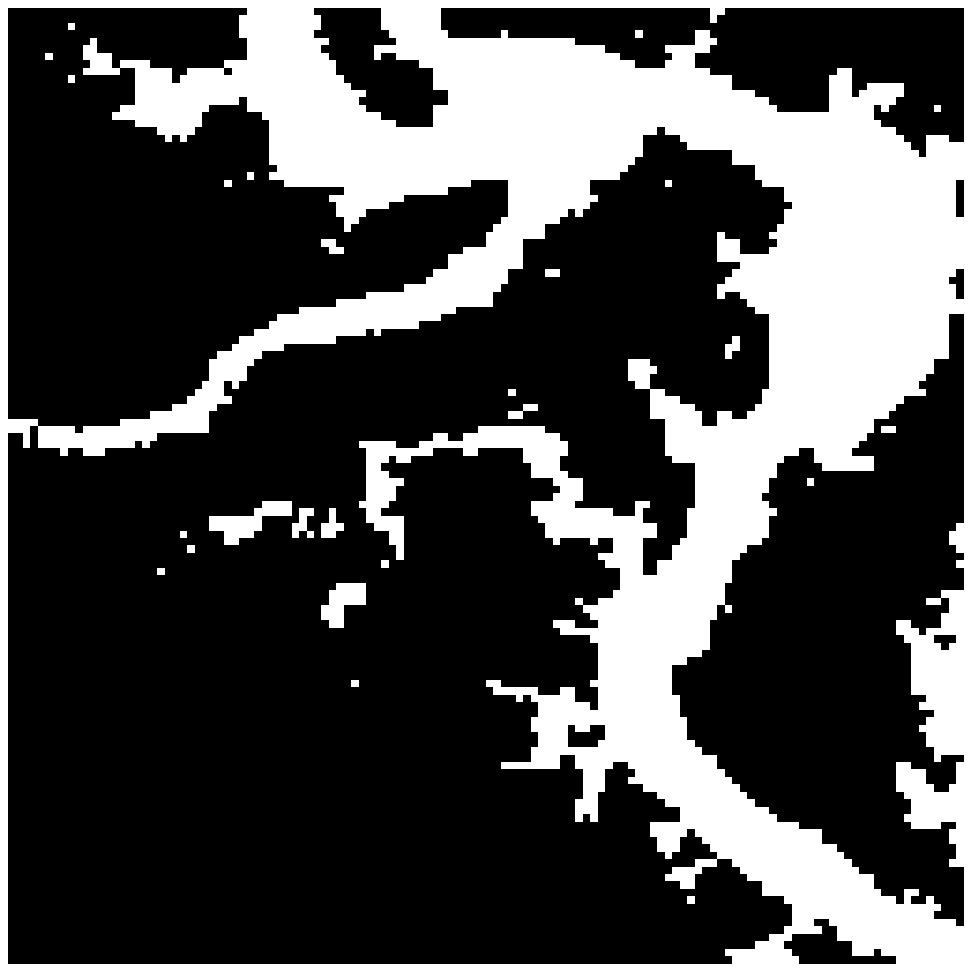,width=0.14\linewidth,clip=} &
\epsfig{file=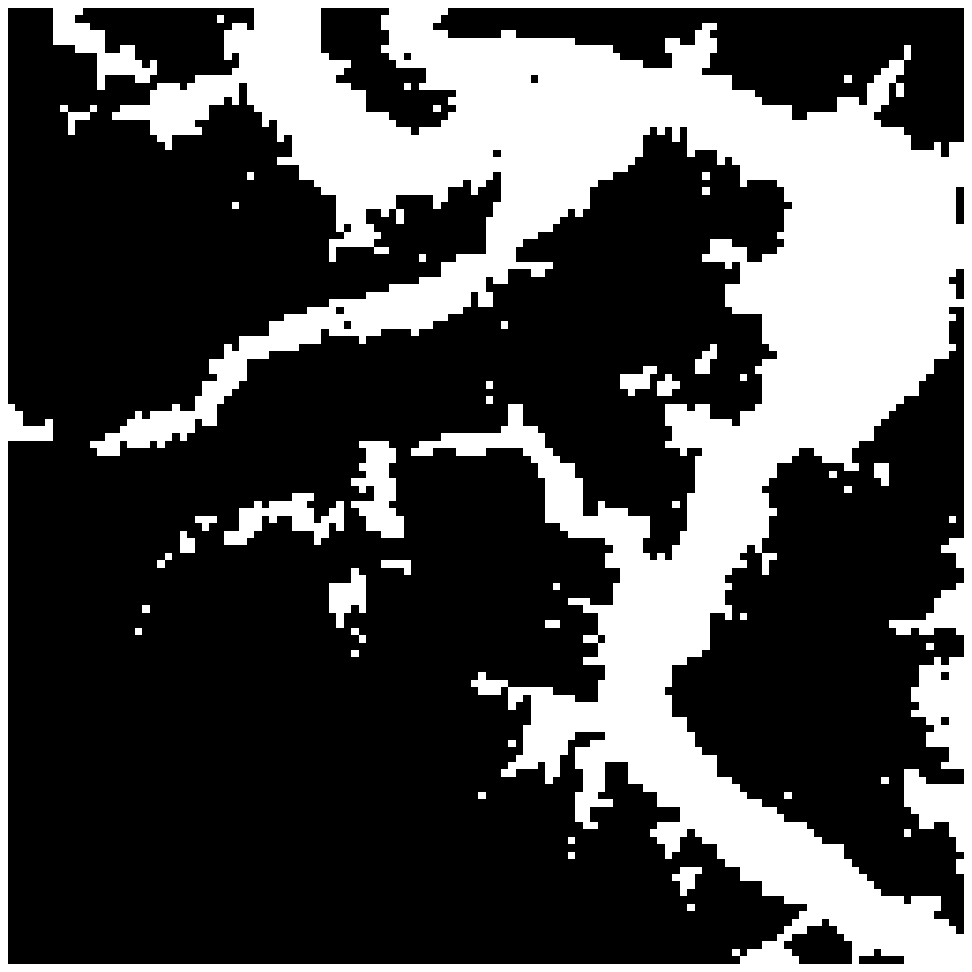,width=0.14\linewidth,clip=} &
\epsfig{file=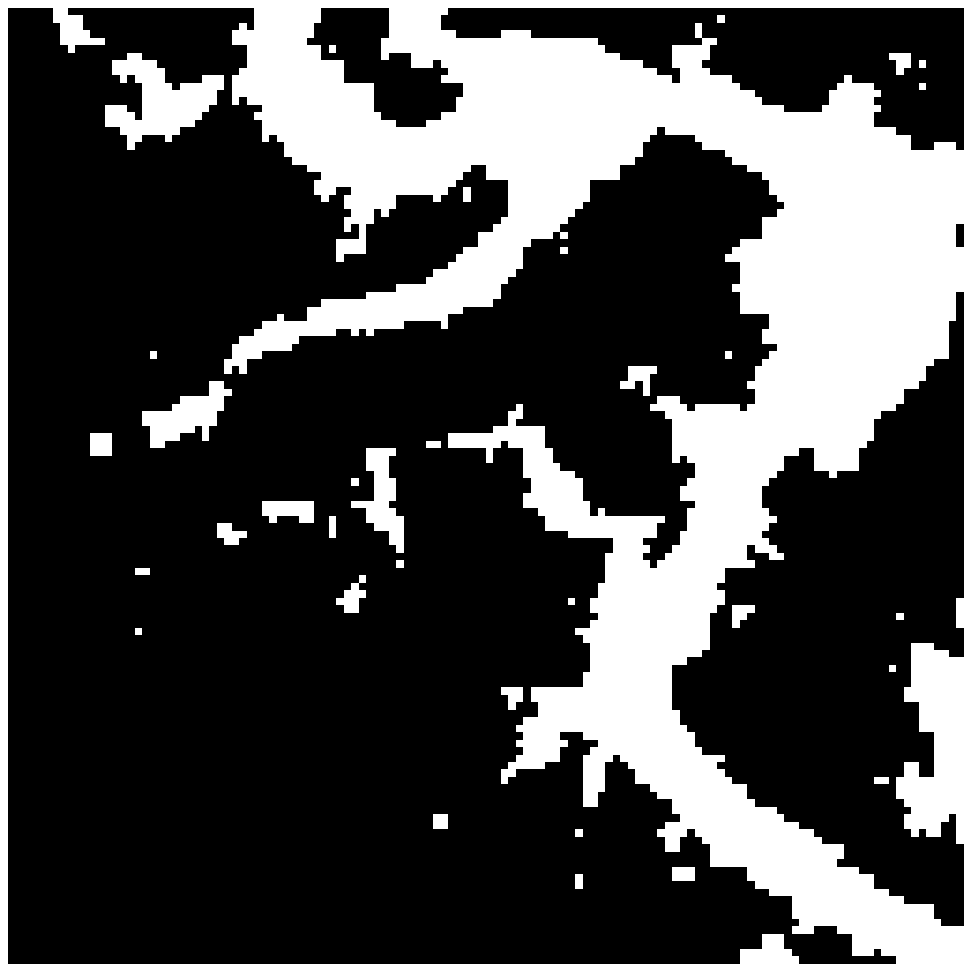,width=0.14\linewidth,clip=} &
\epsfig{file=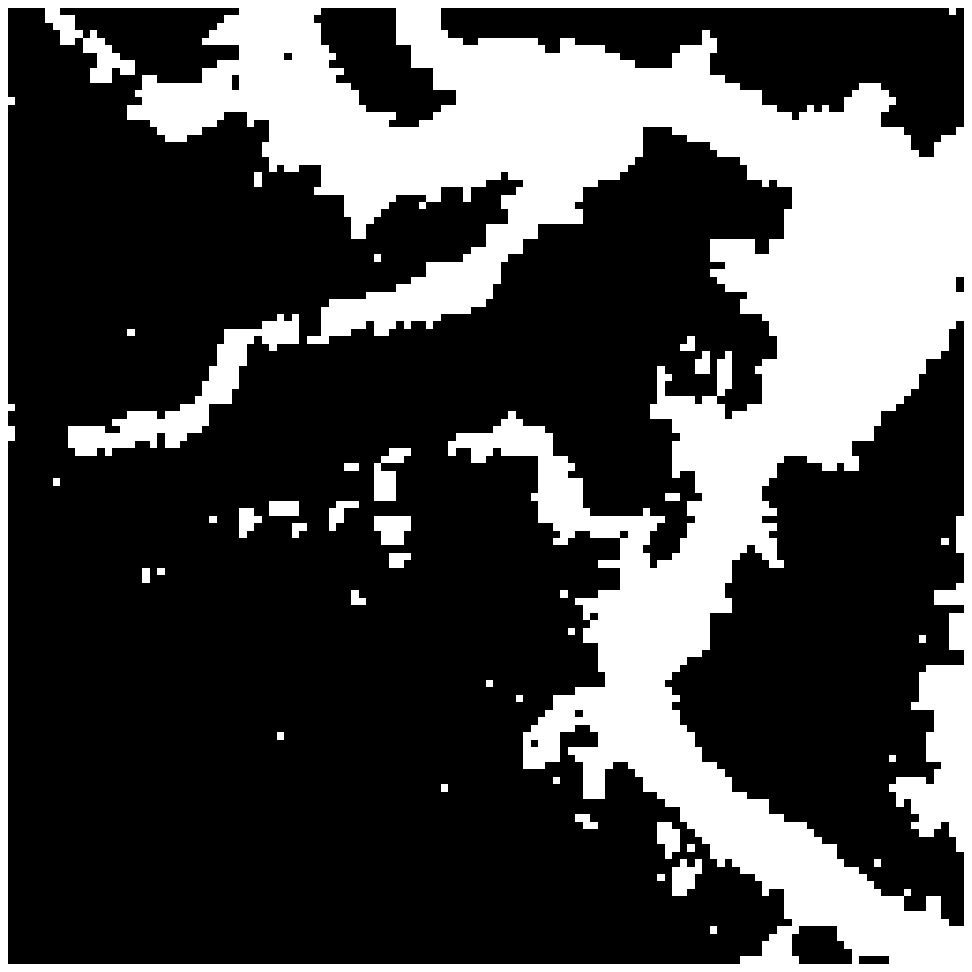,width=0.14\linewidth,clip=} &
\epsfig{file=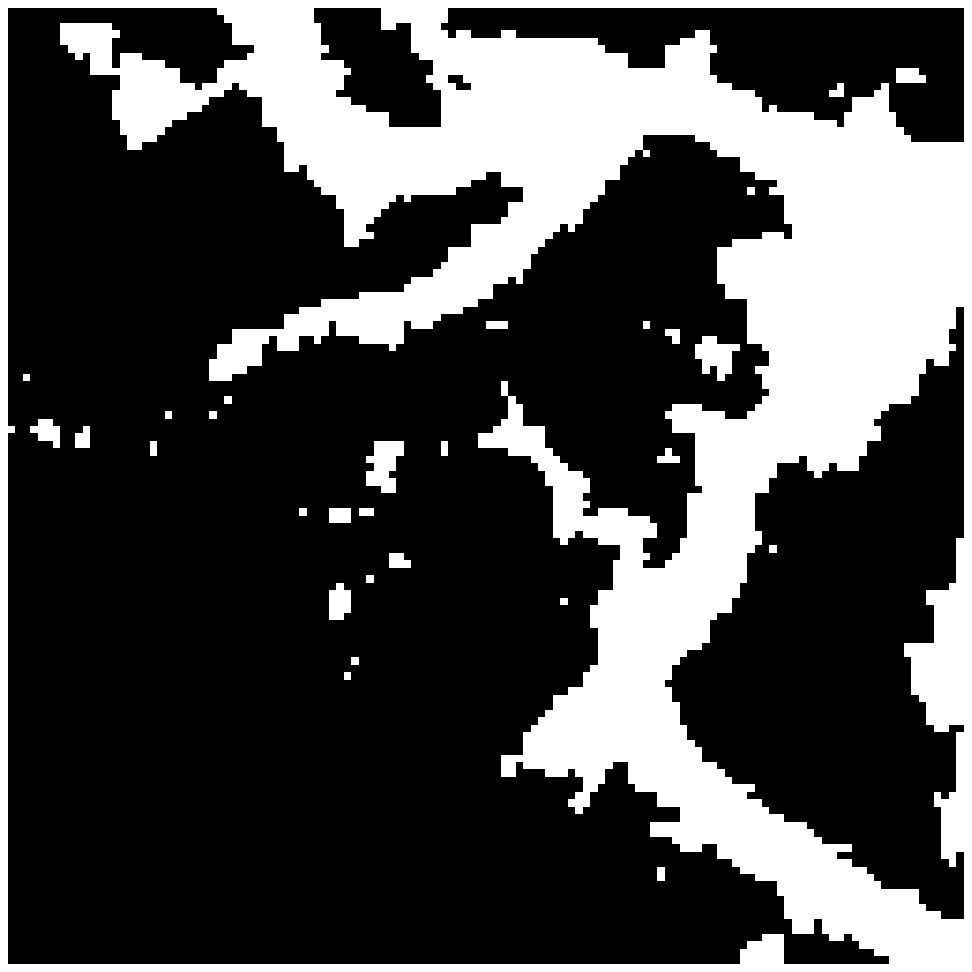,width=0.14\linewidth,clip=} &
\epsfig{file=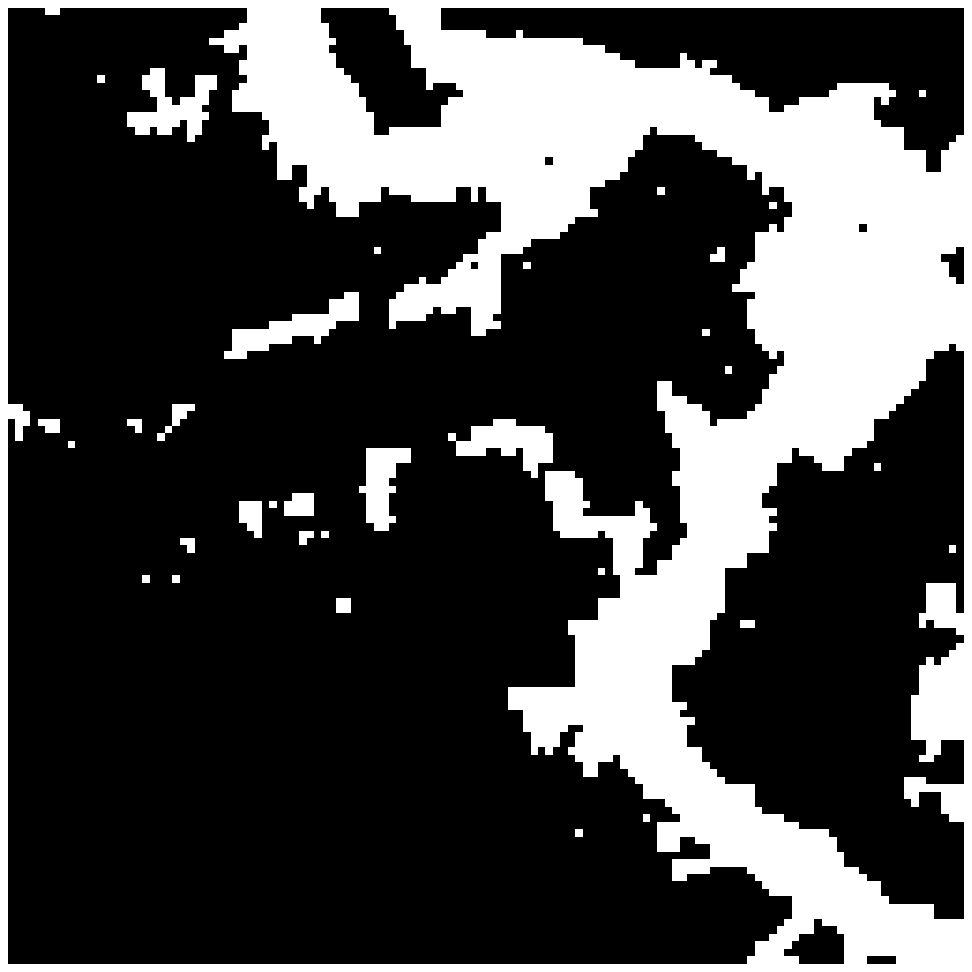,width=0.14\linewidth,clip=}\\
\epsfig{file=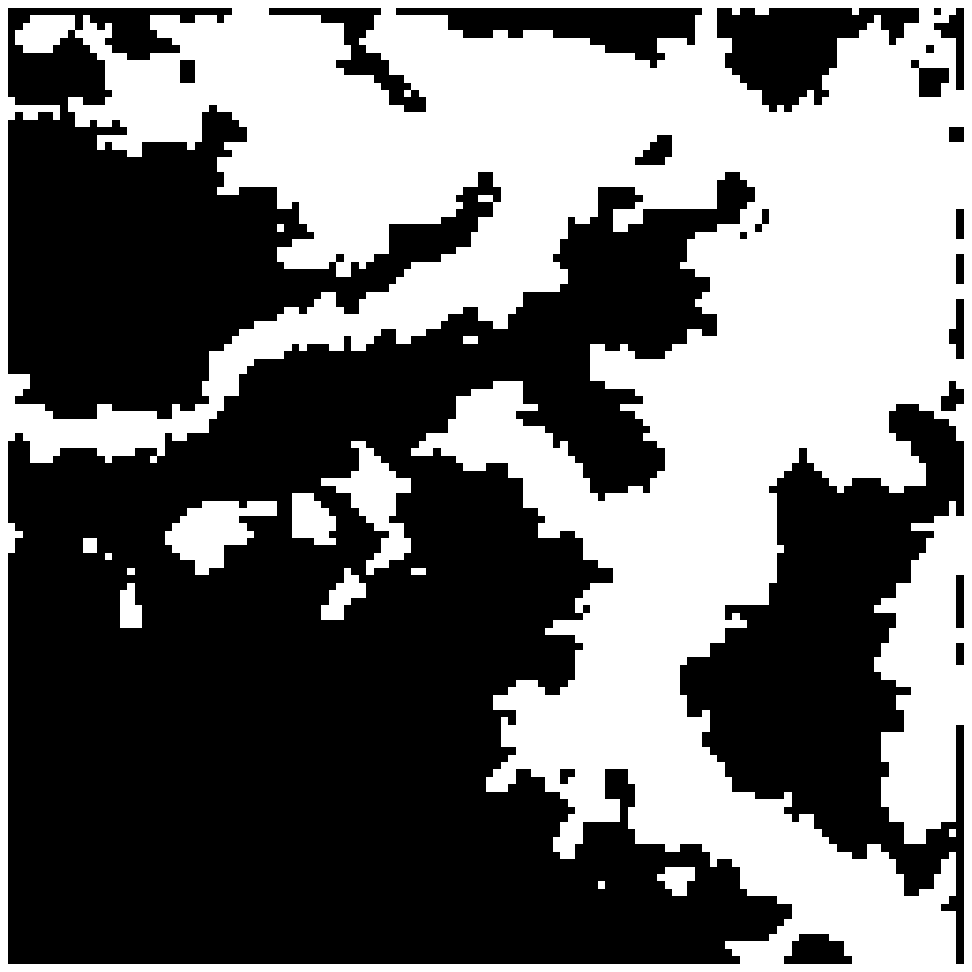,width=0.14\linewidth,clip=} &
\epsfig{file=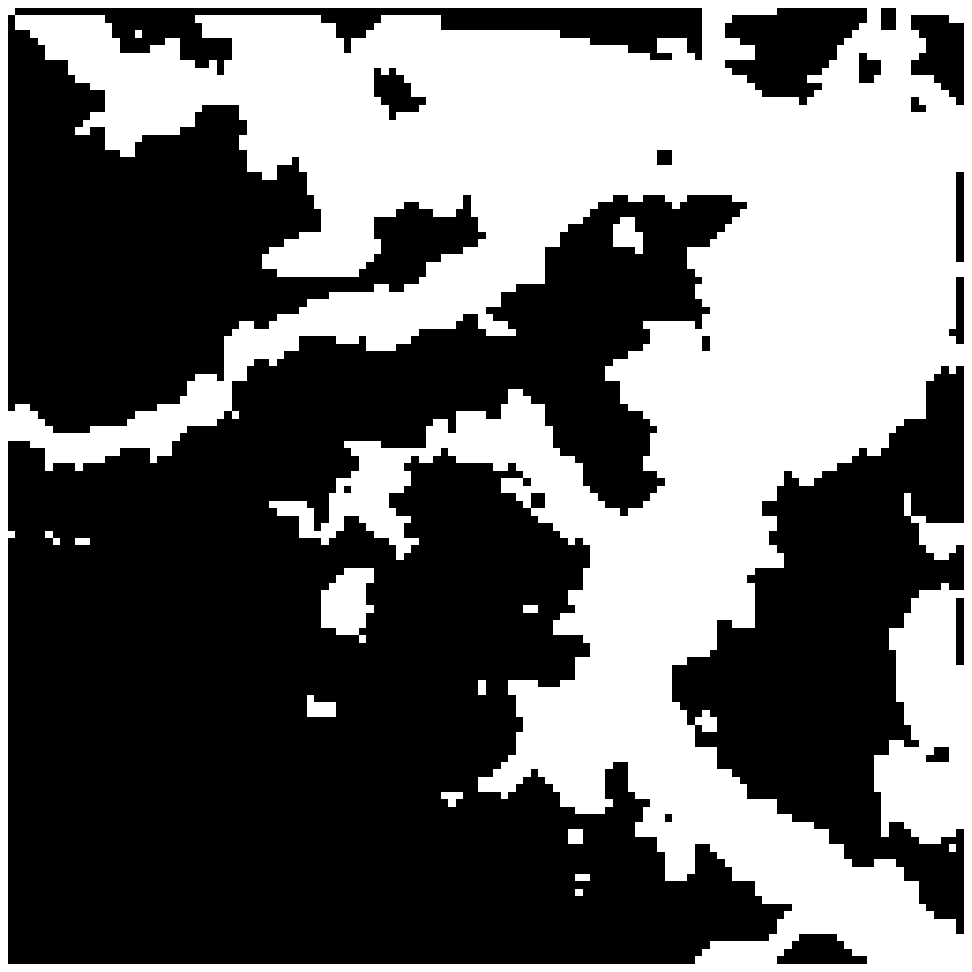,width=0.14\linewidth,clip=} &
\epsfig{file=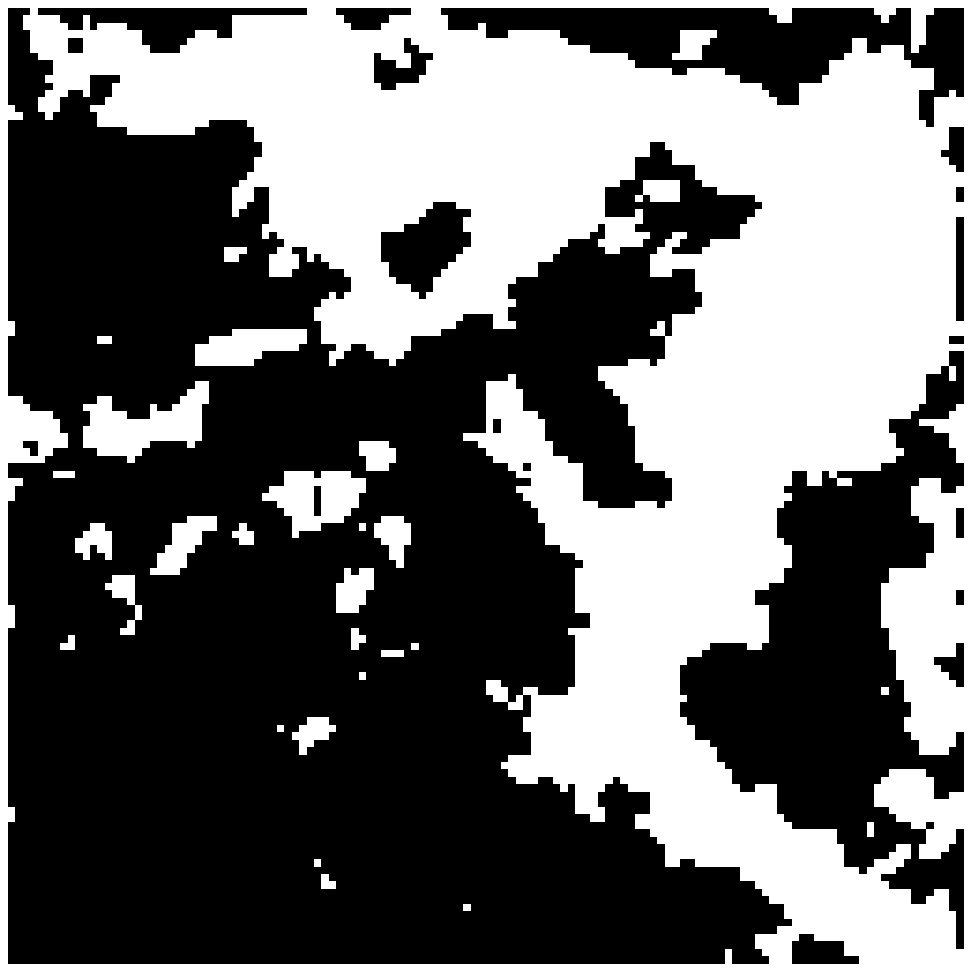,width=0.14\linewidth,clip=} &
\epsfig{file=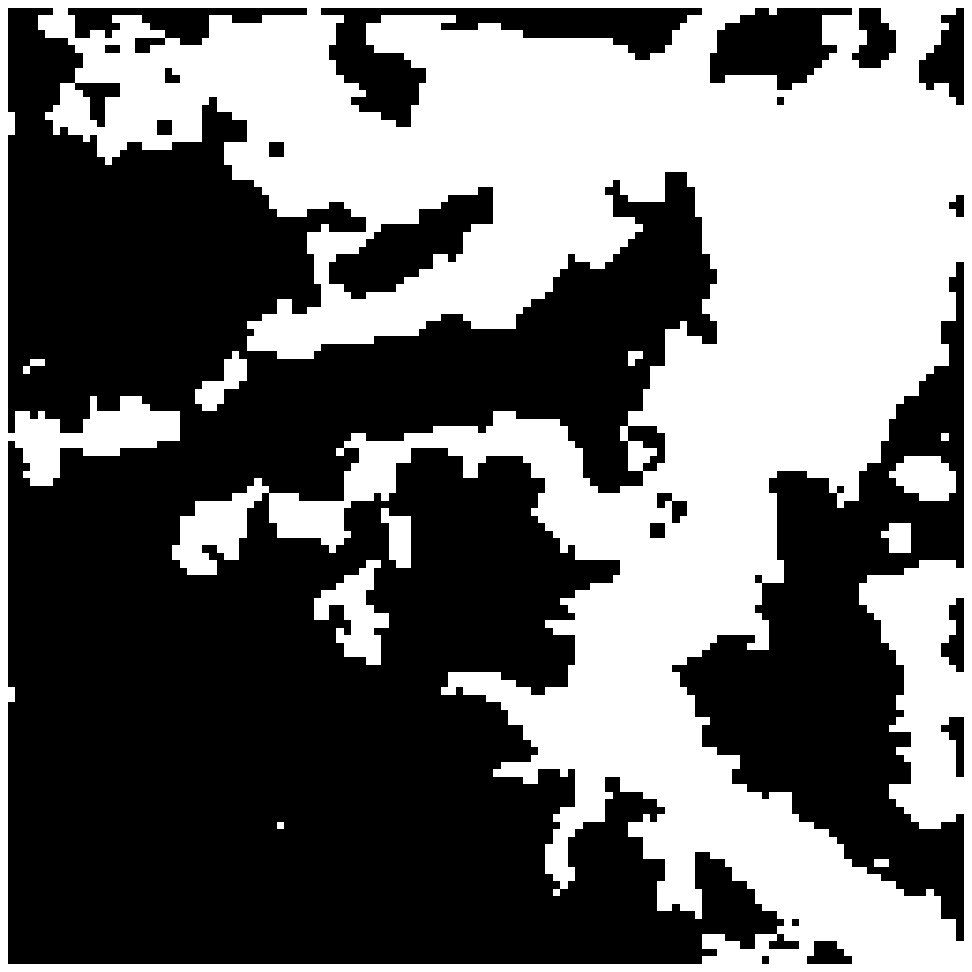,width=0.14\linewidth,clip=} &
\epsfig{file=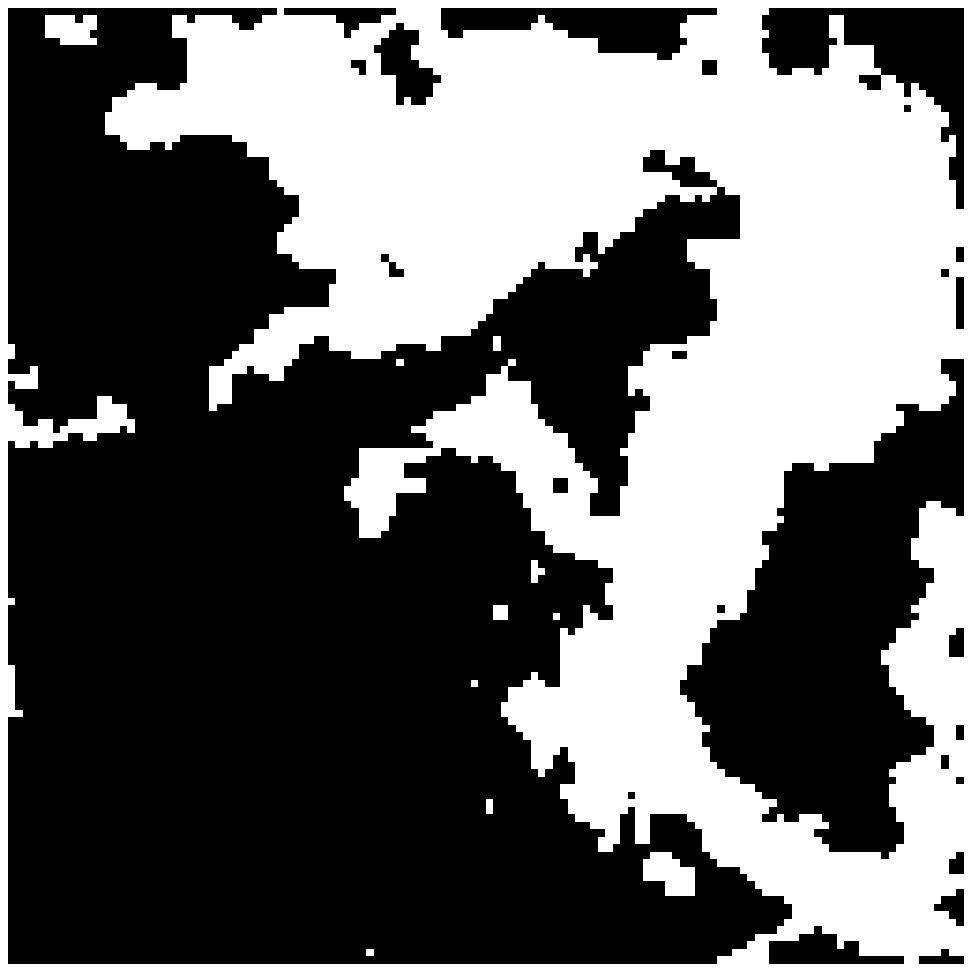,width=0.14\linewidth,clip=} &
\epsfig{file=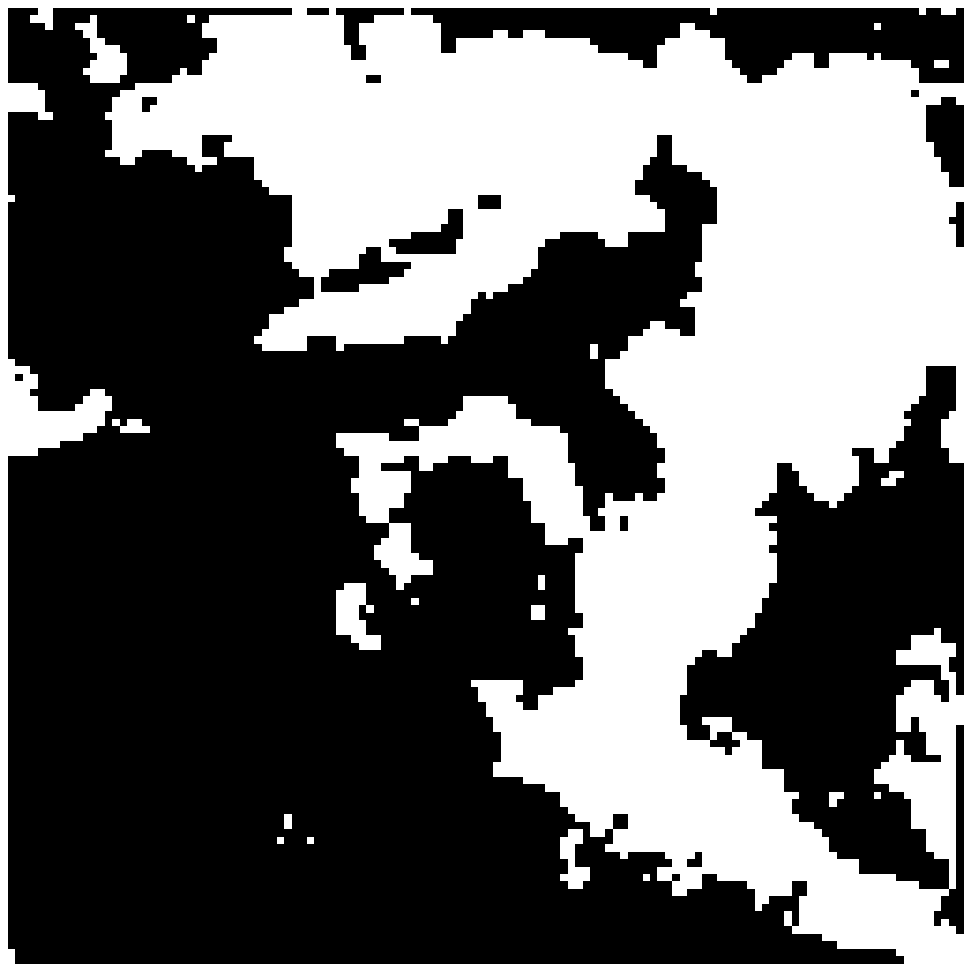,width=0.14\linewidth,clip=}\\
(a) & (b) & (c) & (d) & (e) & (f)\\
\end{tabular}
\end{center}

\caption{Reconstructions of the $\gamma=70$ level set of the St. Louis river image, where row $1$ depicts estimates obtained using our proposed approach and row $2$ depicts the estimates obtained via the approach in \cite{KWR2011}. Columns $(a)$ and $(b)$ correspond to reconstructions from $k=p$ projection measurements with additive noise of $\sigma = 0$ and $\sigma = 10$, respectively. Columns $(c)$ and $(d)$ correspond to reconstructions from $k=p/2$ projection measurements ($50$\% of the total) with additive noise of  $\sigma = 0$ and $\sigma = 10$, respectively. Columns $(e)$ and $(f)$ correspond to reconstructions from $k=p/4$ projection measurements ($25$\% of the total) with additive noise of  $\sigma = 0$ and $\sigma = 10$, respectively.
}
\label{res3}
\end{figure*}

\begin{figure*}[!ht]
\begin{center}
\begin{tabular}{cccccc}
\epsfig{file=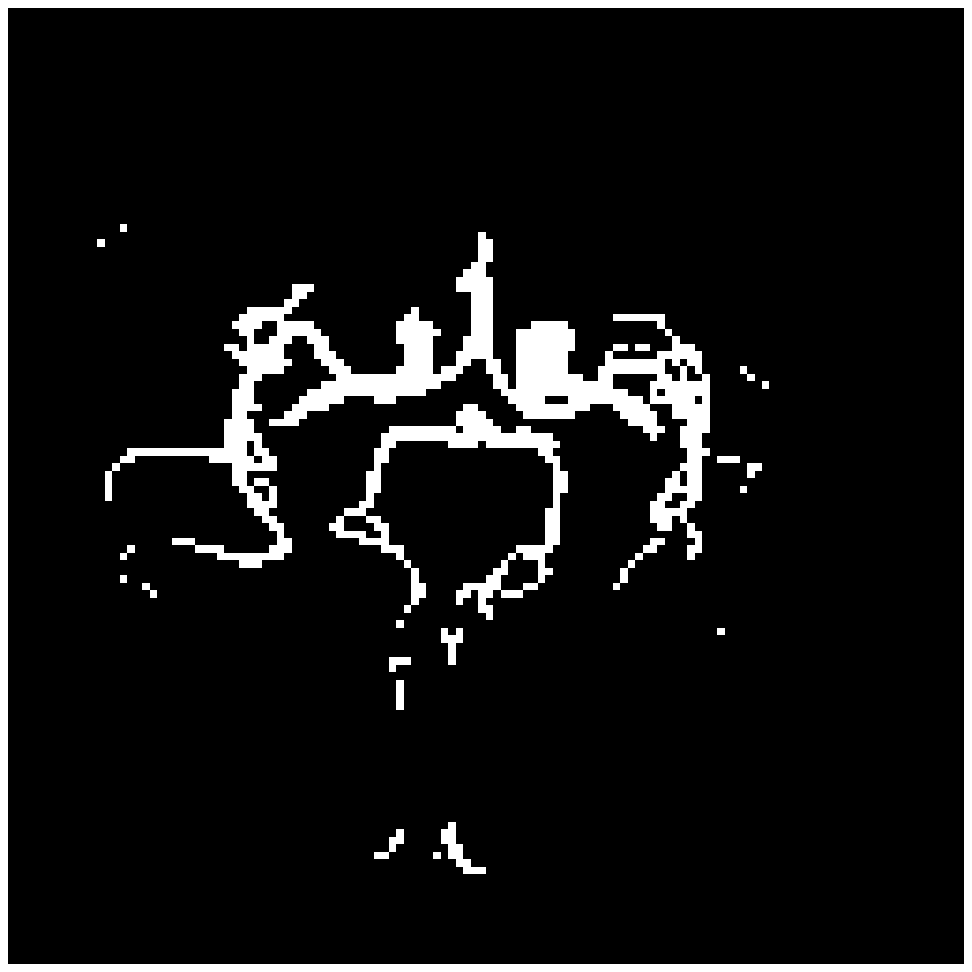,width=0.14\linewidth,clip=} &
\epsfig{file=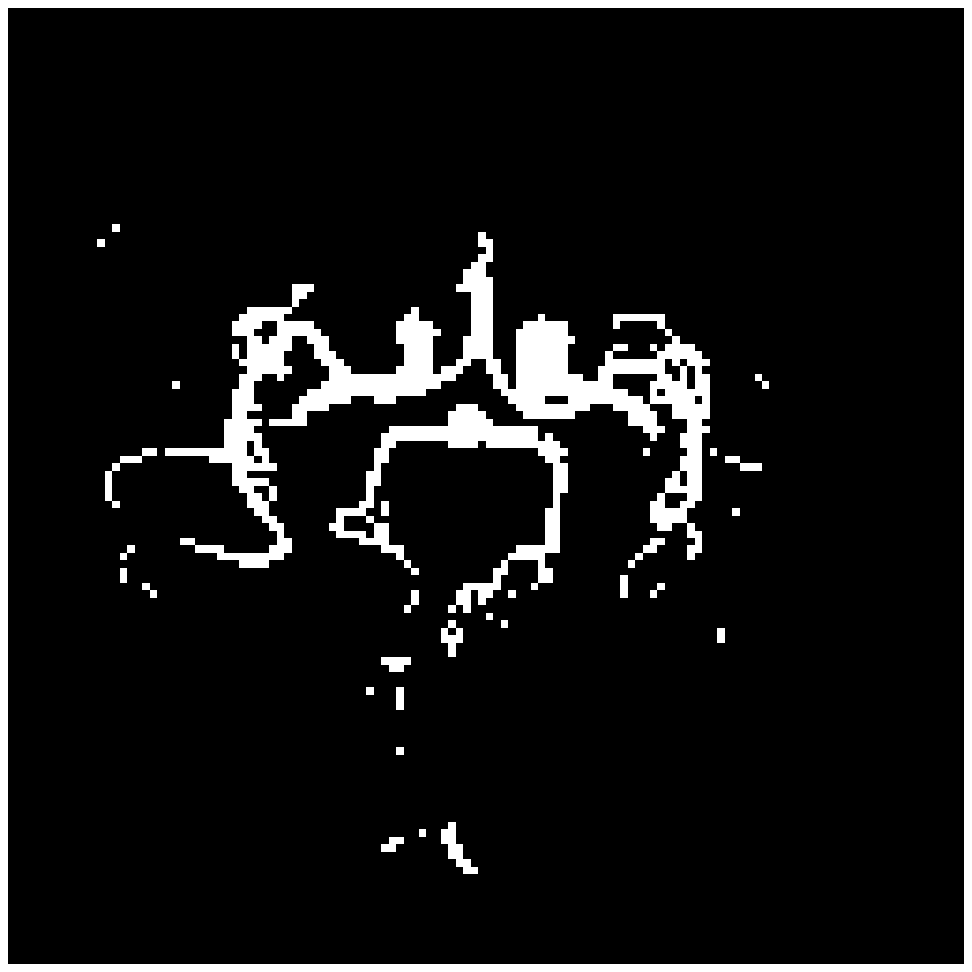,width=0.14\linewidth,clip=} &
\epsfig{file=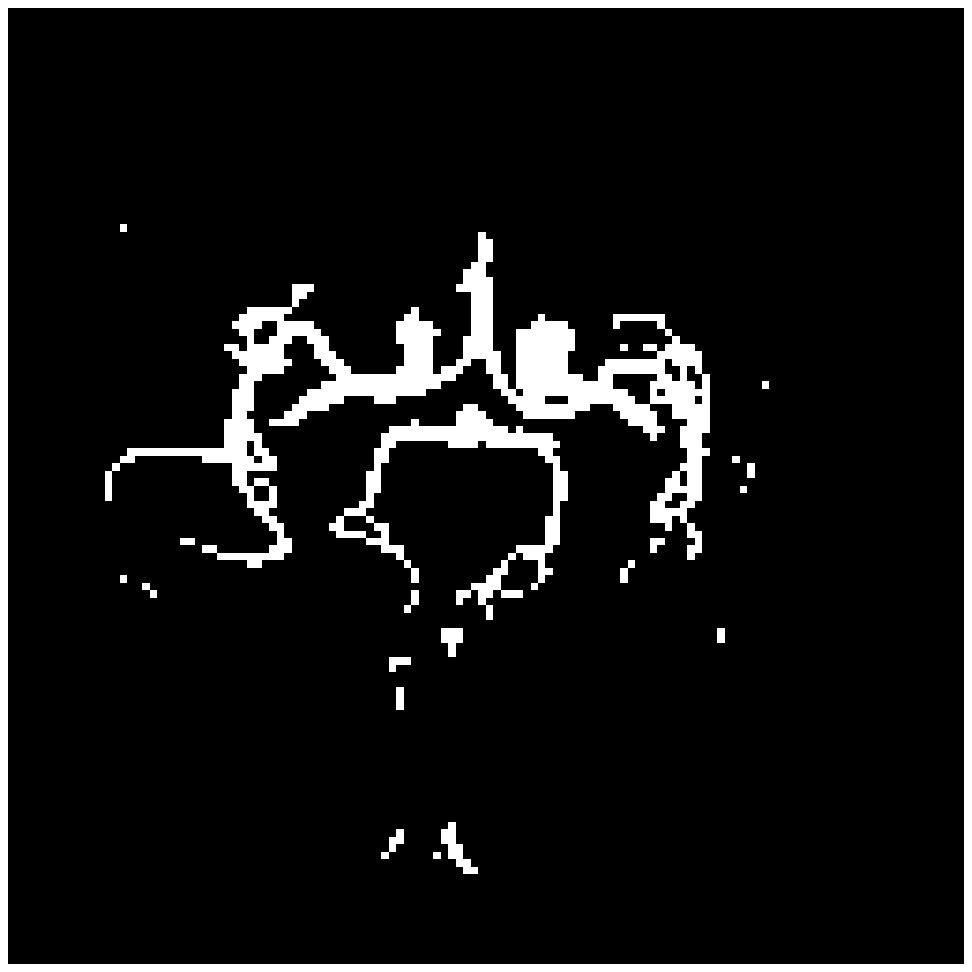,width=0.14\linewidth,clip=} &
\epsfig{file=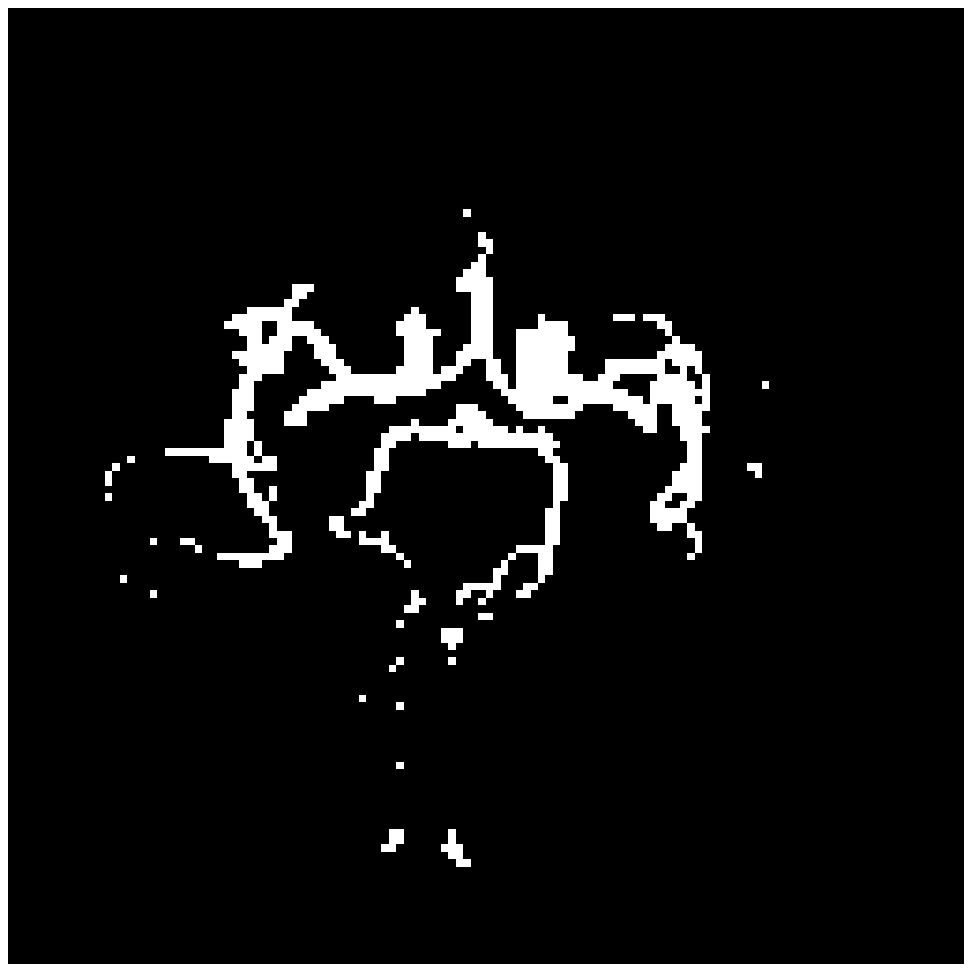,width=0.14\linewidth,clip=} &
\epsfig{file=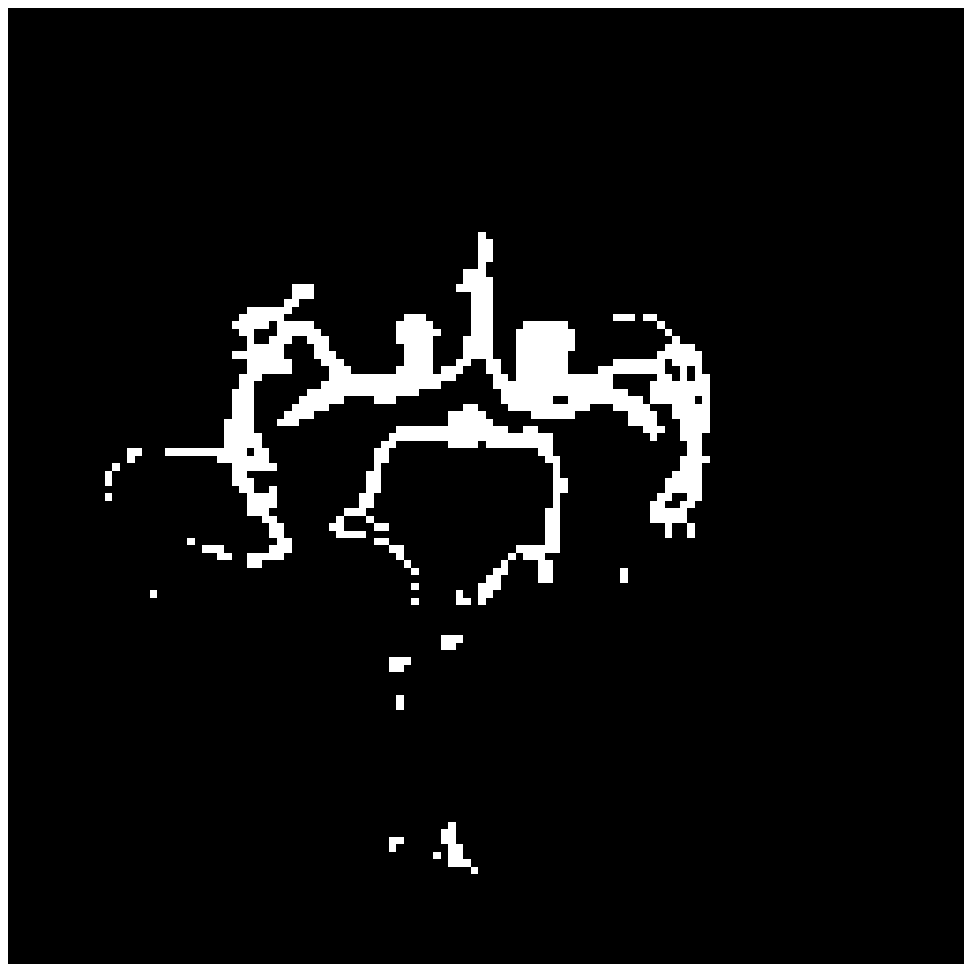,width=0.14\linewidth,clip=} &
\epsfig{file=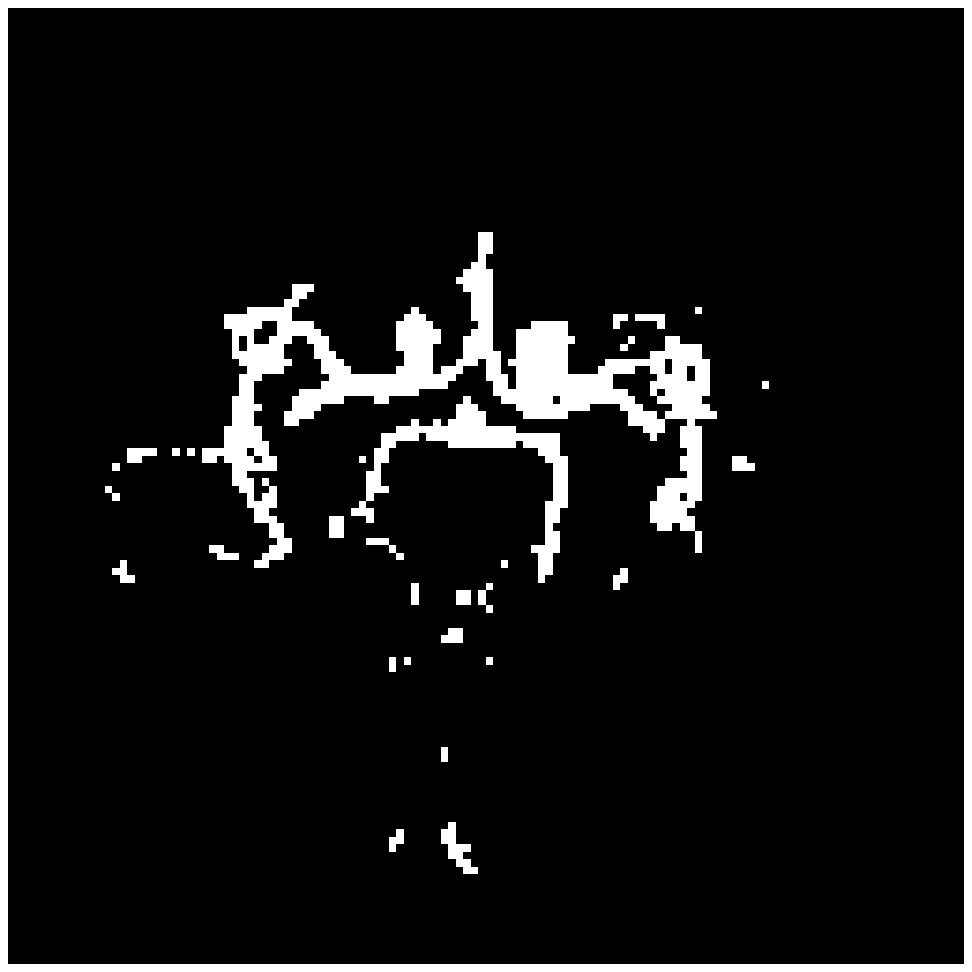,width=0.14\linewidth,clip=}\\

\epsfig{file=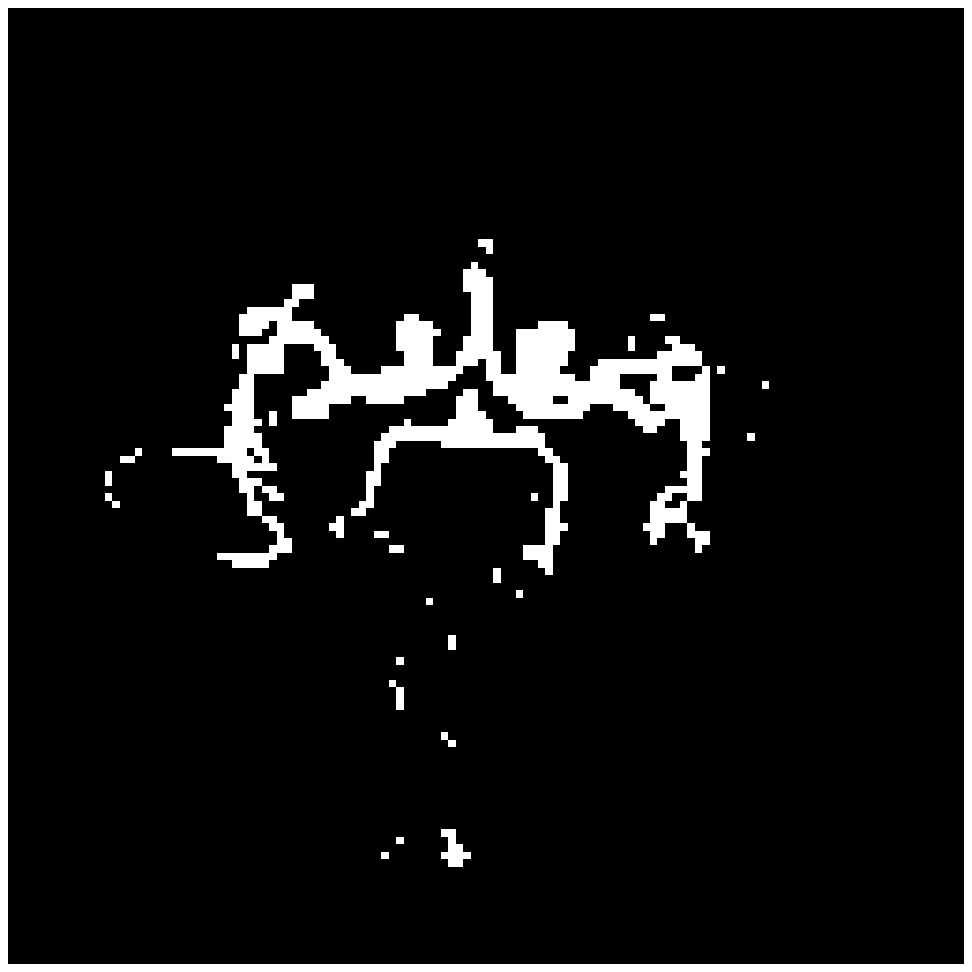,width=0.14\linewidth,clip=} &
\epsfig{file=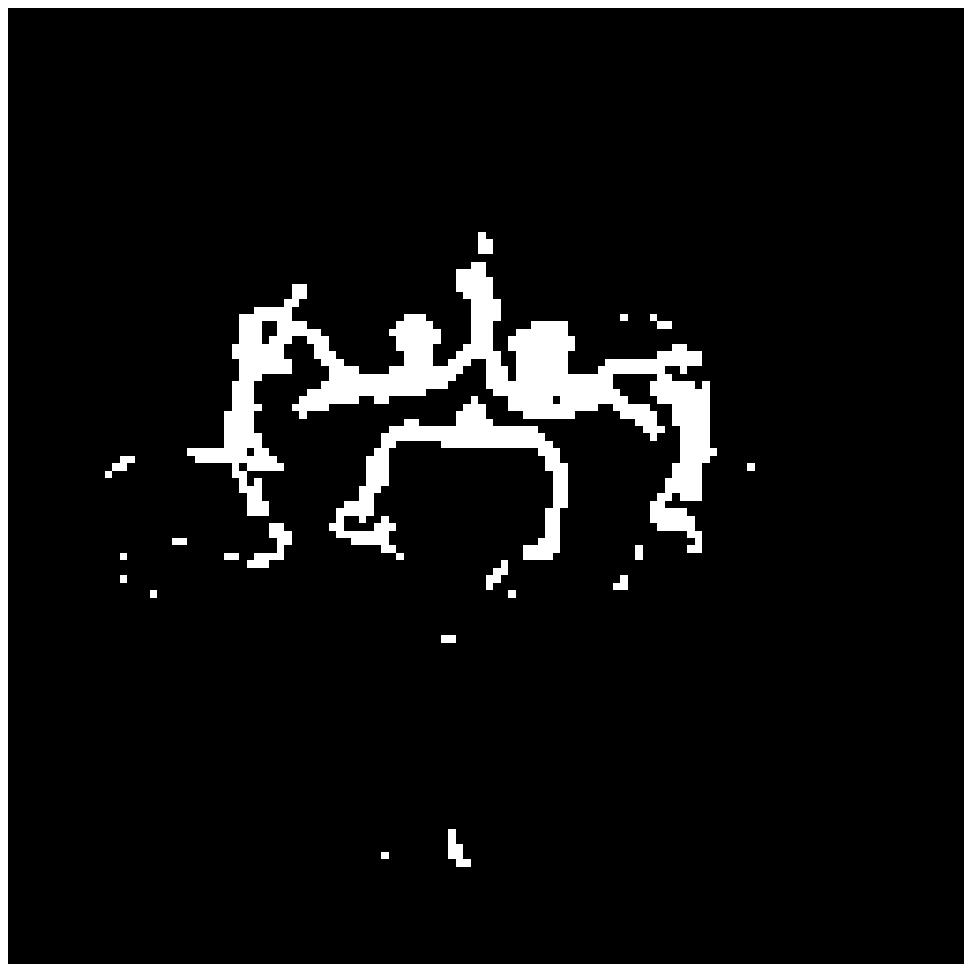,width=0.14\linewidth,clip=} &
\epsfig{file=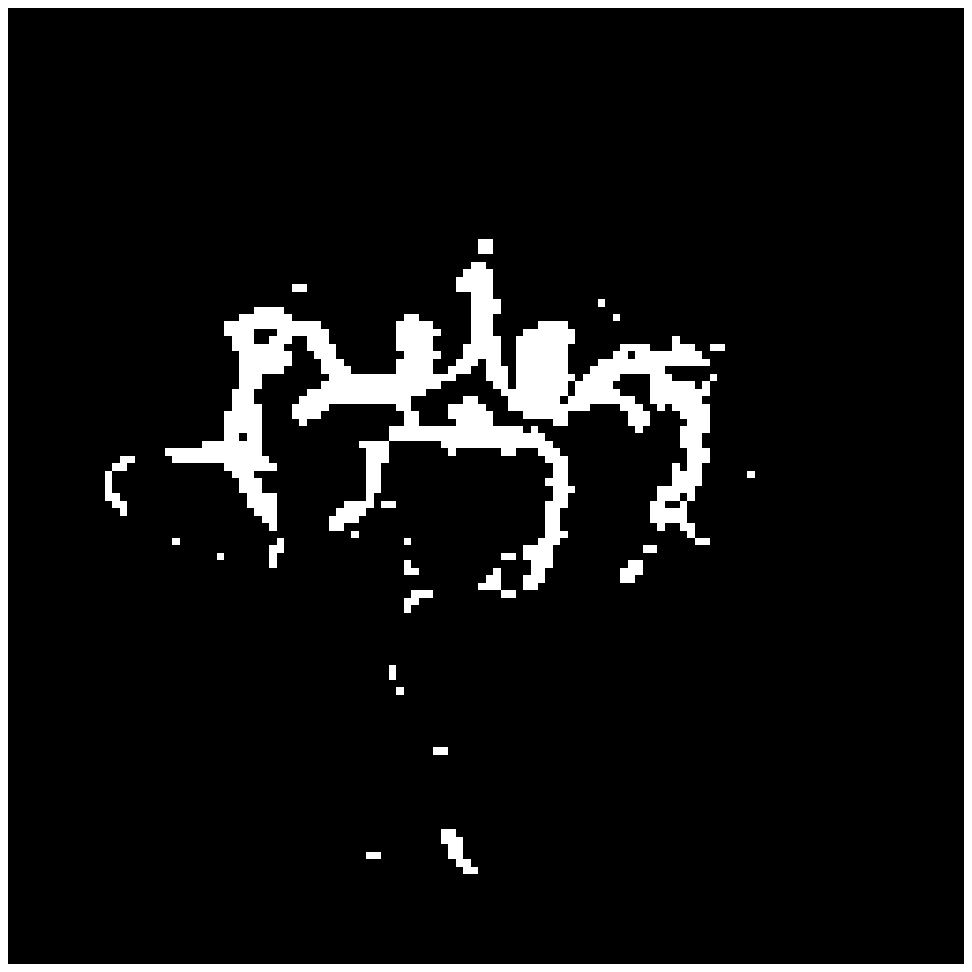,width=0.14\linewidth,clip=} &
\epsfig{file=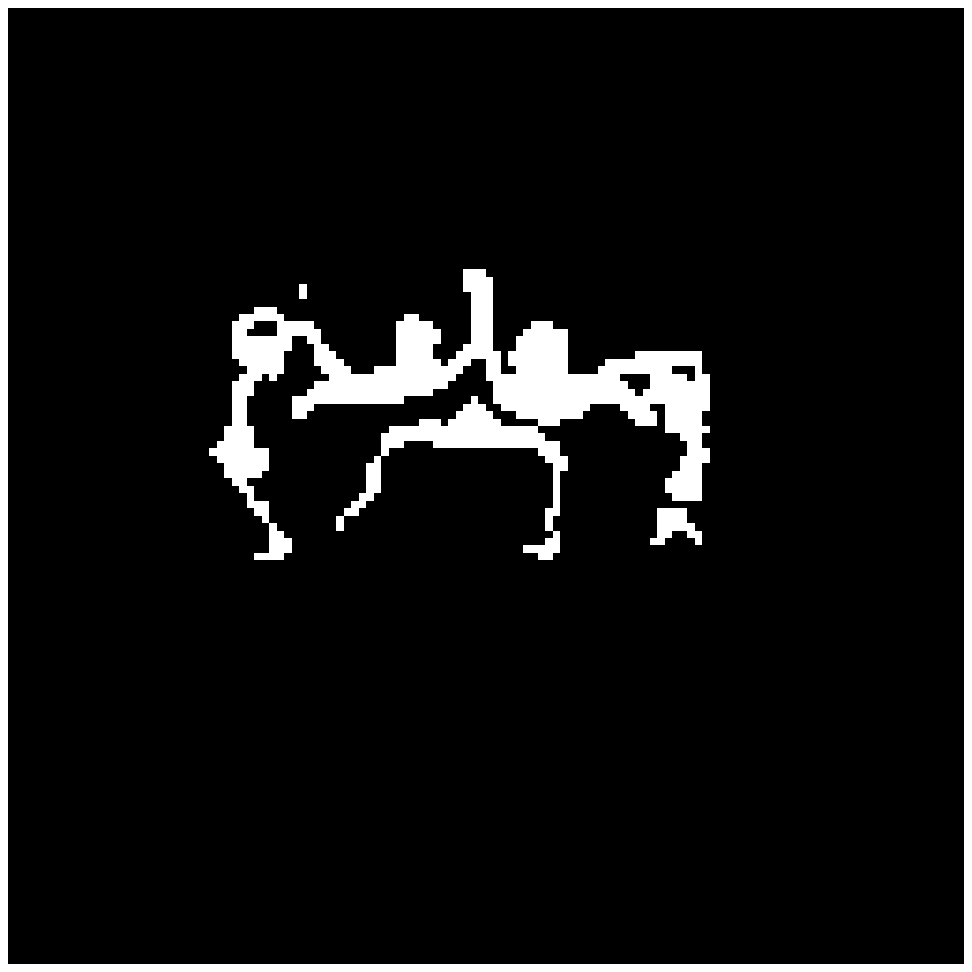,width=0.14\linewidth,clip=} &
\epsfig{file=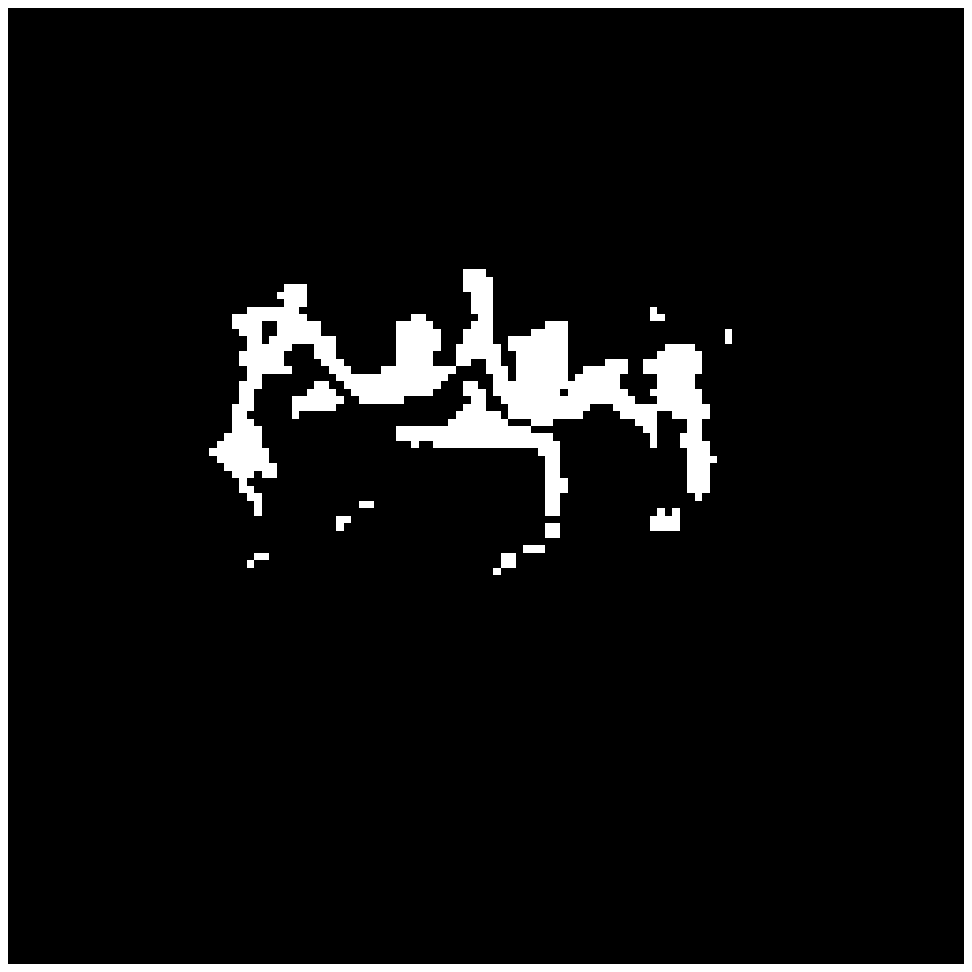,width=0.14\linewidth,clip=} &
\epsfig{file=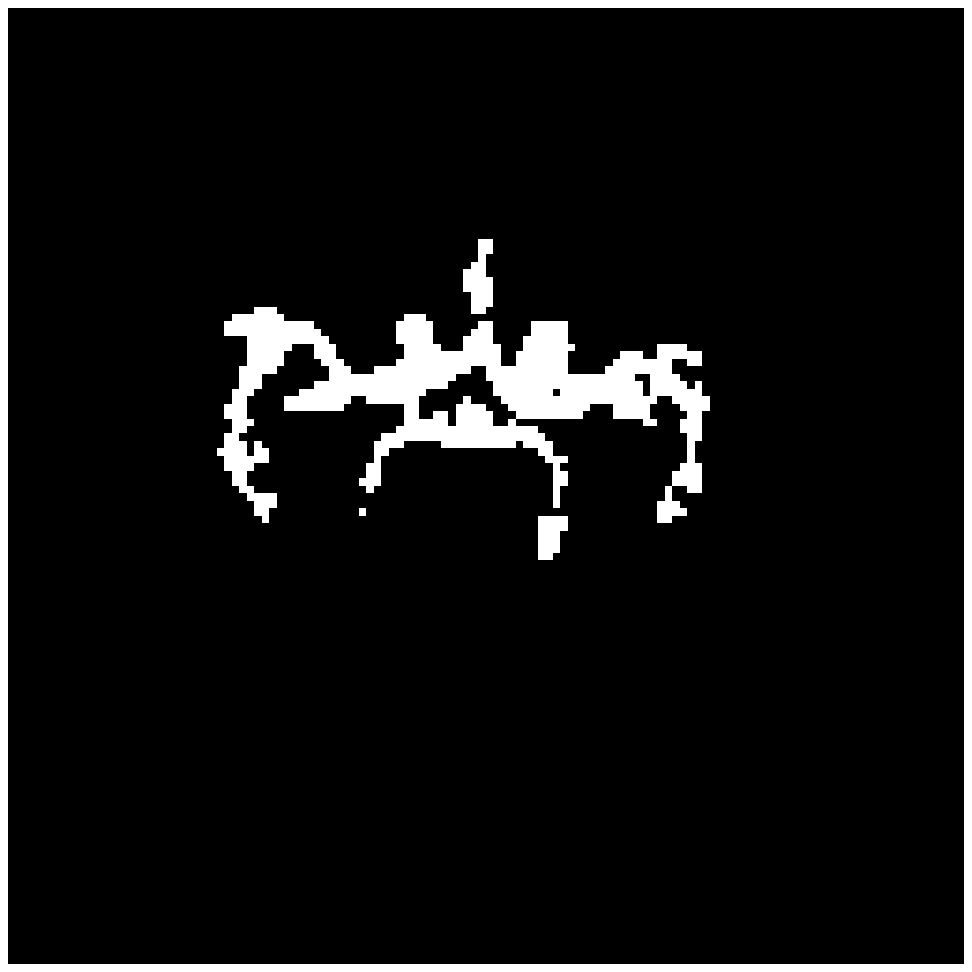,width=0.14\linewidth,clip=}\\
(a) & (b) & (c) & (d) & (e) & (f)\\
\end{tabular}
\end{center}

\caption{Reconstructions of the $\gamma=60$ level set of the magnetic resonance angiography image, where row $1$ depicts estimates obtained using our proposed approach and row $2$ depicts the estimates obtained via the approach in \cite{KWR2011}. Columns $(a)$ and $(b)$ correspond to reconstructions from $k=p$ projection measurements with additive noise of $\sigma = 0$ and $\sigma = 10$, respectively. Columns $(c)$ and $(d)$ correspond to reconstructions from $k=p/2$ projection measurements ($50$\% of the total) with additive noise of $\sigma = 0$ and $\sigma = 10$, respectively. Columns $(e)$ and $(f)$ correspond to reconstructions from $k=p/4$ projection measurements ($25$\% of the total) with additive noise of $\sigma = 0$ and $\sigma = 10$, respectively.
}
\label{res4}
\end{figure*}

Fig.\ref{res2} shows a plot of excess risk vs.~number of projection measurements for estimates of the $\gamma=70$ level set of the image in Fig.~\ref{res1}(a) obtained using the approach described here, the state of the art approach from \cite{KWR2011}, and direct thresholding of proxy observations defined in \eqref{proxy}, in a noise-free setting.  The results show that the approach outlined here achieves a much lower excess risk for the same number of noisy measurements.  Figures \ref{res3} and \ref{res4} provide a visual comparison of our approach with the approach of \cite{KWR2011} when estimating the $\gamma=70$ and $\gamma=60$ level sets, respectively, of two different images.  Estimates obtained using each approach, for several different values of $k$ (number of measurements) and different noise levels are shown. Here, as above, the estimates shown for each case correspond to the clairvoyantly-chosen algorithmic parameter $\alpha$ yielding minimum excess risk in each case.  Note that the TV-based approach proposed here performs fairly well even when the number of measurements obtained are much smaller than the ambient dimension --- see, in particular, Fig.\ref{res4}(e) and (f).    

\section{Conclusions}
\label{conc}
We proposed a simple box-constrained total variation (TV) based optimization approach for level set estimation from compressive measurements.  A fast algorithm based on FISTA was discussed, and simulation results demonstrate the effectiveness of this approach, relative to existing techniques, in the compressive level set estimation problem. Future work in this direction will entail a more exhaustive simulation-based study of this technique for a variety of test images, as well as examination of the performance of this approach from other forms of undersampled data, such as subsampled Fourier data, which arises in tomographic and magnetic resonance imaging applications.    

\bibliographystyle{IEEEtran}
\bibliography{IEEEabrv,refer}

\end{document}